\documentclass{article}
\usepackage{spconf}
\usepackage{amsmath,graphicx}
\usepackage{multirow}
\usepackage{booktabs}
\usepackage[textfont=it,tableposition=top]{caption}
\usepackage{color}

\title{Toward domain-invariant speech recognition via large scale training}
\name{Arun Narayanan, Ananya Misra, Khe Chai Sim, Golan Pundak, Anshuman Tripathi}
\secondlinename{Mohamed Elfeky, Parisa Haghani, Trevor Strohman, Michiel Bacchiani}
\address{Google, USA}
\begin{document}
\maketitle
\begin{abstract}
  Current state-of-the-art automatic speech recognition systems are trained to work in specific `domains', defined based on factors like application, sampling rate and codec. When such recognizers are used in conditions that do not match the training domain, performance significantly drops. This work explores the idea of building a single domain-invariant model for varied use-cases by combining large scale training data from multiple application domains. Our final system is trained using \emph{162,000 hours} of speech. Additionally, each utterance is artificially distorted during training to simulate effects like background noise, codec distortion, and sampling rates. Our results show that, even at such a scale, a model thus trained works almost as well as those fine-tuned to specific subsets: A single model can be robust to multiple application domains, and variations like codecs and noise. More importantly, such models generalize better to unseen conditions and allow for rapid adaptation -- we show that by using as little as \emph{10 hours} of data from a new domain, an adapted domain-invariant model can match performance of a domain-specific model trained from scratch using 70 times as much data. We also highlight some of the limitations of such models and areas that need addressing in future work.
\end{abstract}

\section{Introduction}
\label{sec:intro}
  Automatic speech recognition (ASR) has come a long way in the last few years, with state-of-the-art systems performing close to human performance \cite{Stolcke2017human, Saon2017swbd}. Even so, most ASR systems are trained to work well in highly constrained settings, targeting specific use cases. Such systems perform poorly when used in conditions not seen during training. This {\em mismatch} problem has been widely studied in the context of robustness to background noise and mixed bandwidths  \cite{kim2017mtr, peddinti2016mtr, Yu2013FeatureLearningDNN}. But there has not been a lot of work that addresses other forms of mismatch, e.g., application domains and codecs.

  In this work, we address domain robustness in a more general setting. We broadly use the term `domain' to mean a logical group of utterances that share some common characteristics. Examples include application domains like voice search, video-captioning, call-center, etc., and sub-categories based on other forms of similarities like sample rate, noise, and the codec used to encode a waveform. 

  There is a lot of literature that addresses certain specific aspects of domain invariance. For robustness to noise, multicondition training (MTR) using simulated noisy utterances has been shown to generalize well \cite{kim2017mtr, peddinti2016mtr}. In fact, the gains over MTR with specialized feature enhancement is usually minimal \cite{Narayanan2013DNNSeparation, Yu2013FeatureLearningDNN}. Similarly, mixed bandwidth training has been shown to handle multiple sample rates simultaneously, without any need for explicit reconstruction of missing bands \cite{Yu2013FeatureLearningDNN}. In the context of multiple application domains, training by pooling data was shown to work well for domains with limited resources in \cite{ghahremani17-transfer}.

  Unlike existing studies that address only some form of domain robustness, like noise, the presented work scales it up to simultaneously address several aspects of domain invariance: Robustness to a variety of application domains while operating at multiple sampling rates using multiple codecs for encoding input, and in the presence of background noise. To build such a \emph{multidomain} model, we pool training data from several sources, and simulate conditions like background noise, codecs and sample rates. Since mixed-bandwidths and noise robustness have received a lot of attention in the literature, we will focus more on less explored areas of robustness like application domains and codecs. 
  
  We present results using, to the best of our knowledge, the largest speech database ever used to train a single model -- \emph{162,000 hours} of speech before simulating additional conditions like noise, codec and sampling rates. After including simulated distortions, the probability that the model sees the same utterance in the same mixing condition twice during training is close to 0, which implies that the final size of speech material seen during training is 162,000 hours $\times$ the number of epochs during training. Surprisingly, even though the multidomain model is trained by combining diverse datasets, it works almost as well as the domain-specific models. This is despite the fact we did not try to explicitly balance the amount of training data the model sees in each domain during training. It also generalizes better to unseen domains. Most interestingly, we show that such a model can be rapidly adapted to new conditions with very little data. On a previously unseen domain, we get large gains by adapting the multidomain model using as little as 10 hours of data, outperforming a model trained only on the new domain using 700 hours speech. Our results also show what domains, or combination of domains, are harder to model at this scale and warrant more research.

  The rest of the paper is organized as follows. {Sec.~\ref{sec:prior}} discusses prior work. The models and training strategies used in this work are described in {Sec.~\ref{sec:desc}}. Experimental settings and results are presented in {Sec.~\ref{sec:exp}}. We conclude in {Sec.~\ref{sec:concl}}.

\section{Prior work}\label{sec:prior}
  There have been a number of studies that address invariance to background noise. Training on noisy data and using simulated noisy utterances to augment clean training are widely used \cite{Yu2013FeatureLearningDNN, kim2017mtr, peddinti2016mtr, sainath2017multichannel}. Specialized techniques like masking \cite{Narayanan2013IRM} and beamforming \cite{yoshioka2015ntt} (in the case of multi-microphone input) help in certain conditions. Adaptation has also been used to address noise. In \cite{mirsamadi2017multi}, a model is adapted to a previously unseen far-field condition by learning a linear transform of an intermediate layer, where the noise or the domain information is best encoded. Paired clean and noisy unlabeled data and model-distillation were used in \cite{Li2017domain} to adapt a clean model to noisy conditions. Our work differs from these studies in that noise mismatch is only one of the dimensions we explore. Furthermore, our goal is to train a model that works well in multiple conditions, not just noise.
  
  Similar to noise, mixed bandwidth training helps generalize to multiple sampling rates \cite{Yu2013FeatureLearningDNN}. In \cite{Yu2013FeatureLearningDNN}, the authors train the acoustic model with data sampled at both {16~kHz} and {8~kHz}. When computing features for {8~kHz} input, the high frequency logmel bands are set to zeros.  While more sophisticated techniques like reconstructing high frequency bands have been proposed \cite{gao2016experimental}, the gains over mixed bandwidth training are often small.
  
  Multiple application domains have typically been studied in the context of transfer learning \cite{bengio2012transfer}. In \cite{ghahremani17-transfer}, transfer learning is used to adapt a model trained on switchboard \cite{switchboard} to improve performance on domains with smaller training sets like WSJ \cite{Paul1992} and AMI \cite{ami-corpus}. They also show that pooling data from multiple low-resource domains work better than transfer learning. Unlike \cite{ghahremani17-transfer}, the current work studies domain robustness in a much larger scale, where data sparsity is not necessarily a challenge. We also study other forms of mismatch like codec, and consider many more applications domains.

  Domain adaptation has been widely studied in the machine learning and vision literature. A typical formulation is to learn a model for a target domain with limited unlabeled data, using supervised data from a source domain. Domain adversarial training and variants are widely used for this \cite{ganin2016domain, tzeng2017adversarial}. 

\section{Model description}\label{sec:desc}
\begin{figure}[ht]
  \centering
  \includegraphics[width=1.0\linewidth]{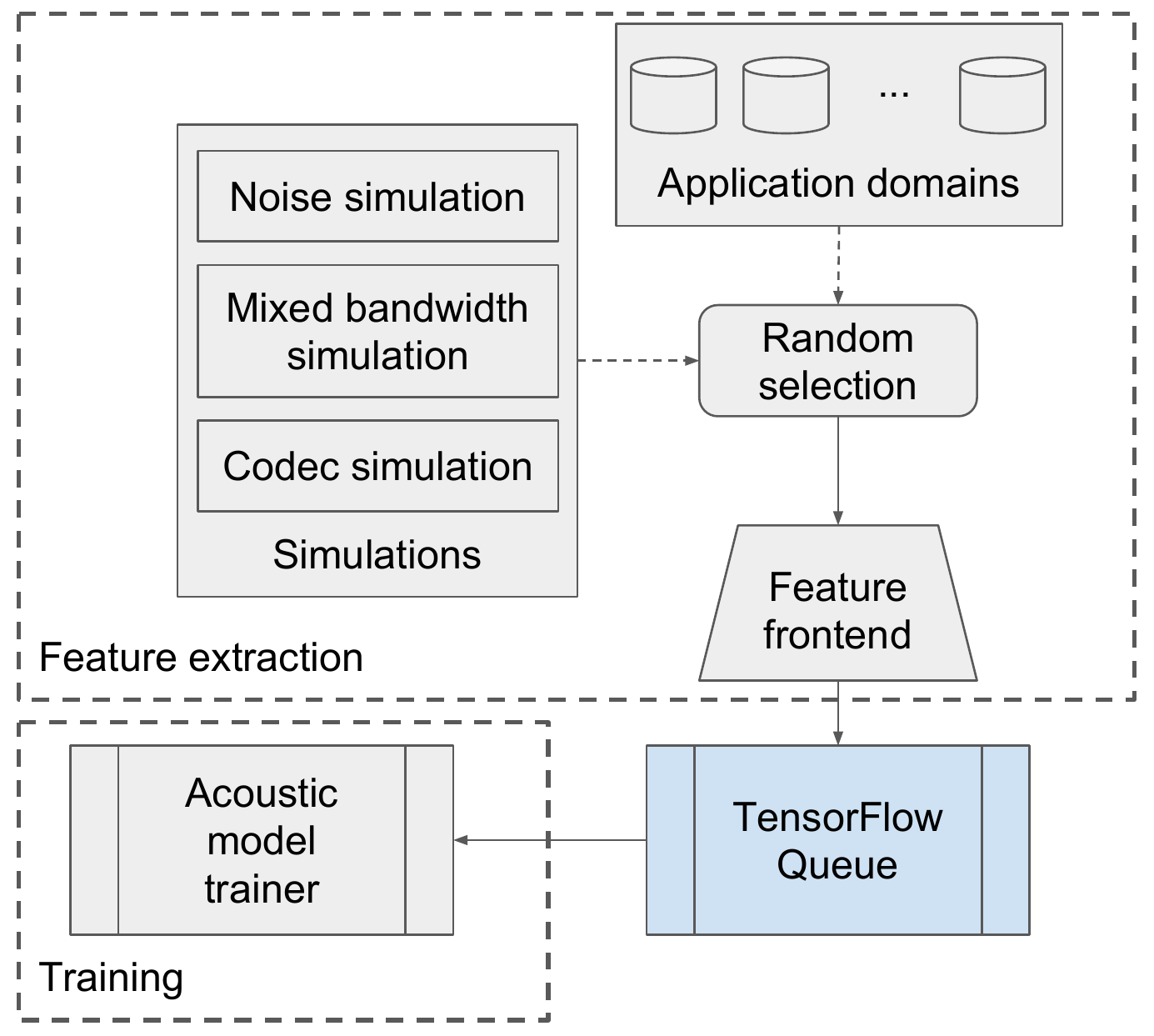}
  \caption{Block diagram for multidomain training.}
  \label{fig:md_training}
\end{figure}
 {Fig.~\ref{fig:md_training}} shows a block diagram for the processing pathways. To generate the training set, we pool data from multiple application domains. Utterances are chosen randomly from the pooled set during training. Given an utterance, we randomly apply zero or more simulated perturbations. This includes 1) noise simulation via a room simulator, 2) changing the sample rate, and 3) encoding and decoding using a lossy codec. The features extracted from the resulting utterance is pushed into a queue. The model reads from the queue during training. Feature extraction and training happens asynchronously to prevent the feature computation overhead from slowing down training. The stages are described in more detail below. 
 
\subsection{Noise simulation}\label{subsec:noise}
  For noise simulation, we use a setting similar to the one used in \cite{kim2017mtr} that has been shown to work well for noisy and far-field voice search tasks. During training, a noise configuration, which defines mixing conditions like the size of the room, reveberation time, position of the microphone, speech and noise sources, signal to noise ratio (SNR), etc., for each training utterance is randomly sampled from a collection of 3 million pre-generated configurations. A simulated room may contain 0 to 4 noise sources, which is mixed with speech at an SNR between 0 to 30 dB. The reverberation time is set to be between 0 and 900 msec, with a target to mic distance between 1 and 10 meters (see \cite{kim2017mtr} for details). The noise snippets used for simulation come from a collection of YouTube, cafeteria, and real-life noises. We expect these snippets to cover noise conditions encountered in typical use cases like voice-search on mobile phones.
 
\subsection{Mixed bandwidth simulation}
  We only consider {8~kHz} and {16~kHz} sample rates in this work, since all the data we have use these rates. Note that for sample rates greater than {16~kHz}, the input can be downsampled without any loss of information since our feature frontend only uses frequencies up to {7.5~kHz}. The majority of the data that we use for training is sampled at {16~kHz}. To balance the training set, we randomly downsample an utterance to {8~kHz} with a probability of 0.5. Our feature extraction frontend is configured to operate at {16~kHz}, so the waveform is then upsampled back to {16~kHz} before feature extraction. Since we use logmel features as input to the acoustic model (see Sec.~\ref{subsec:feature}), this is very similar to adding zeros to the high frequency logmel bands when the input is at {8~kHz} \cite{Yu2013FeatureLearningDNN}.

\subsection{Codec simulation}
  To simulate a variety of audio encodings before transmission to the recognition system, we encode and decode each waveform with a randomly selected codec. We train with the MPEG-2 Audio Layer 3 (MP3) and Advanced Audio Coding (AAC) codecs, both of which perform perceptual audio coding~\cite{brandenburg1999mp3}. We apply these at constant bit rate using the implementations in FFmpeg~\cite{ffmpeg}. The set of 7 codec conditions sampled uniformly at random for training consists of: MP3 at bit rates of 128, 32 and 23 kbps, AAC at 128, 64 and 23 kbps, and no codec. Note that the training data we use may already be encoded in arbitrary ways before we obtain it, in which case it is decoded and then re-encoded and re-decoded with the selected codec.
 
\subsection{Feature extraction}\label{subsec:feature}
  The acoustic models are trained using globally normalized logmel features. Input utterances are framed using a 32 msec window, with 10 msec overlap between neighboring frames. 128 dimension logmel features are then extracted, spanning frequencies from 125 Hz to {7.5~kHz}. Four contiguous frames are stacked to form a 512 dimensional feature that is used as input by the acoustic model. Input features are also subsampled by a factor of 3; the acoustic model operates at 33 Hz \cite{pundak2016lfr}.
 
\subsection{Acoustic model}
  We use low frame rate (LFR) models \cite{pundak2016lfr} for acoustic modeling. The acoustic model (AM) predicts tied, context-dependent, single-state phones (CDPhones) every 30 msec. We use 8192 CDPhones; the tree used for state-tying is generated using a subset of the application domains. The AM is a 5 layer unidirectional LSTM \cite{sak2014LSTM}, with 1024 cells in the first 4 layers, and 768 cells in the last layer. The models are cross-entropy (CE) trained, using alignments generated using the original, unperturbed utterances. 
  
\subsection{Language model}
  The focus of the current work is domain invariance of acoustic models, so we use the same language model (LM) in all experiments. The LM is a Bayesian interpolated 5-gram model trained from a variety of data sources like anonymized and aggregated search queries and dictated texts \cite{allauzen2011bayesian}. It consists of around 100 million n-grams and a vocabulary size of 4 million words. The LM is used for single pass decoding.
  
\subsection{Large scale training}
  One of the challenges in training at such a scale is the amount of resources needed, both in terms of disk space and compute. Since we use several types of simulated perturbations, the amount of disk space needed to store features can quickly grow. To deal with this, we only store the original waveforms and their corresponding alignments on disk. Feature are computed on-the-fly during training, after distorting the waveforms based on zero or more perturbations. Some of the perturbations, like simulating a noise condition, can be computationally expensive. But we make use of the fact this can be done asynchronously with training and utilize the input queuing mechanism in TensorFlow \cite{abadi2016tensorflow} to store the computed features. The trainer reads from the queue, and is not affected by the slowness of feature computation process as long as there are enough jobs feeding the queue \cite{variani2017end}. Feature computation runs on CPUs; the models are trained on GPUs using asynchronous stochastic gradient descent. Using 32 GPU workers, each with Nvidia K80 GPUs, one epoch of CE training takes approximately 1.8 days for the multidomain training set. Training converges within 15 -- 20 epochs. It is possible to further speed up training using faster hardware like Nvidia P100 GPUs or TPUs \cite{jouppi2017datacenter}.

\section{Experiments and results}\label{sec:exp}

\subsection{Datasets}\label{subsec:datasets}

\begin{table}[ht]
  \caption{Data distribution for various application domains.}
  \label{tab:data_dist}
  \centering
  \begin{tabular}{cc}
    \toprule
    \textbf{Application} & \textbf{Dataset size} \\
    \textbf{Domain} & \textbf{(approx. hours)}  \\
    \midrule
    Voicesearch  & $16$k \\
    Dictation    & $18$k \\
    Other search & $1$k \\ 
    Farfield     & $8$k \\
    Call-center  & $1$k \\
    YouTube      & $117$k \\
    \midrule
    Total        & $162$k \\
    \bottomrule
  \end{tabular}
\end{table}

\begin{figure}[ht]
  \centering
  \includegraphics[width=1.0\linewidth]{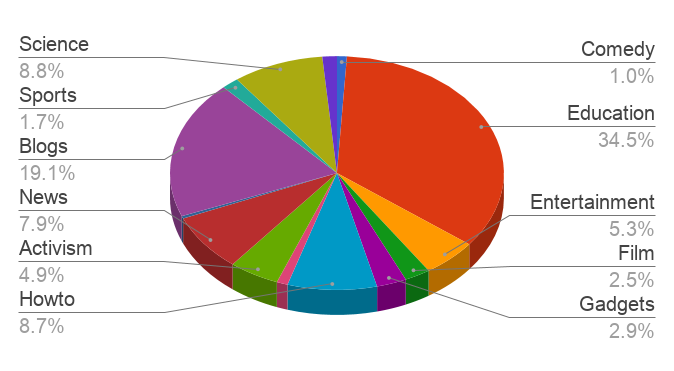}
  \caption{Distribution of data in the YouTube training set.}
  \label{fig:yt_dist}
\end{figure}

  We experiment using a number of training sets that cover a variety of applications, all in English. This includes: \emph{Voicesearch}, \emph{Dictation}, \emph{Other search}, \emph{Farfield} (Google Home), \emph{YouTube} (video-captioning), and \emph{Call-center}. The amount of data available for each of these application domains, shown in {Tab.~\ref{tab:data_dist}}, is quite different, which is typical in any large scale multidomain setting. As can be seen, the training set is dominated by \emph{YouTube} -- around 70\%. But the \emph{YouTube} set is quite diverse and includes multiple sub-domains. {Fig.~\ref{fig:yt_dist}} shows the distribution of various application domains included in the \emph{YouTube} training set. 
  
  The evaluation sets come from similar domains as training -- \emph{Voicesearch}, \emph{Dictation}, \emph{Call-center}, and \emph{YouTube} -- with additional variations to account for other forms of mismatch. Since voicesearch data originating from mobile phone is most likely to be corrupted by noise and codec\footnote{Data compression is used in low bandwidth conditions, which is typical for mobile phones not connected to broadband.}, we perturb the \emph{Voicesearch} test set with these distortions. The noisy sets are constructed using MTR configurations and noise segments not used in training. The \emph{Dictation} set is downsampled to {8~kHz} since that is another common use-case. A \emph{Telephony} test set is used to evaluate out-of-domain performance. It is acoustically most similar to \emph{Call-center}, but is more conversational. An analysis of the quantitative similarities of the different sets is presented in the following section. 
  
  All of the sets used for training and evaluation are anonymized, and are representative of Google's voice search, captioning and cloud traffic. The \emph{YouTube} set was transcribed in a semi-supervised fashion \cite{liao2013large,soltau2016neural}; the rest of the datasets are hand-transcribed.
  
\subsection{Dataset analysis}
  
  To understand the data landscape, we apply internal clustering metrics to gauge how well the data is clustered by domain and to estimate similarities between clusters.
  
  We extract 32-dimensional i-vectors, which capture the acoustic characteristics of an utterance~\cite{Senior2014}.
  We compute the silhouette coefficient~\cite{rousseeuw1987} for each point as follows:
  \begin{equation}
  s(i) = \frac{b(i) - a(i)}{\max\{a(i), b(i)\}},
  \end{equation}
  where $s(i)$ is the silhouette for the $i$-th data point $X_i$, $a(i)$ is the mean Euclidean distance of $X_i$ to other points in the same cluster, and $b(i)$ is the mean Euclidean distance of $X_i$ to the nearest neighboring cluster. The higher the silhouette score, the better the point sits within its designated cluster. The silhouette averaged over a cluster thus suggests how well-defined the cluster is; over the entire data set, it quantifies how distinct the clusters are on average. 
  To understand how similar each application domain is to the held-out \emph{Telephony} domain, we perform pairwise silhouette analysis. We also compute a cluster similarity measure $R_{ij}$~\cite{Davies1979}:
  \begin{equation}
  R_{ij} = \frac{S_i + S_j}{M_{ij}},
  \end{equation}
  where $S_i$ is a dispersion measure of cluster $i$ and $M_{ij}$ is a distance measure between clusters $i$ and $j$. Here we define $S_i$ as the average Euclidean distance of points in cluster $i$ to the cluster centroid, and $M_{ij}$ as the Euclidean distance between the centroids of clusters $i$ and $j$. Results from $50$ examples sampled from each application domain (Tab.~\ref{tab:pairwise_clustering}) reinforce that \emph{Telephony} is close to \emph{Call-center} and suggest it is most distinct from \emph{Voicesearch}.
  
\begin{table}[ht]
  \caption{Pairwise clustering metrics between each application domain and the Telephony domain. Lower silhouette scores and higher cluster similarity indicate more overlap.}
  \label{tab:pairwise_clustering}
  \centering
  \begin{tabular}{lrr}
    \toprule
    \textbf{Domain} & \textbf{Average silhouette} & \textbf{Cluster similarity} \\
    \midrule
    Voicesearch & $0.0732$ & $3.30$ \\
    Dictation & $0.0440$ & $4.16$ \\
    Farfield & $0.0443$ & $4.14$ \\
    Call-center & $0.0248$ & $5.22$ \\
    YouTube & $0.0321$ & $4.81$ \\
    \bottomrule
  \end{tabular}
\end{table}

\subsection{Results}\label{subsec:results}
  We first present results when using simulated perturbations. We train on a subset of the application domains for ease of running experiments, and for evaluating the effect of each simulation technique before combining them. Finally, we present results when training with all of the domains and show generalization results to the unseen domain. Mixed bandwidth simulation is used in all experiments. Since training using multiple sample rates has already been shown to work well in prior work \cite{Yu2013FeatureLearningDNN, ghahremani17-transfer}, we did not explore it in detail in the current study.

\subsubsection{Noise}
  To evaluate noise simulation, we train using \emph{Voicesearch}, \emph{Dictation} and \emph{Other search} sets, with and without simulated background noise. Results are shown in {Tab.~\ref{tab:noise}}. As shown, training with simulated noisy data using the MTR settings described in {Sec.~\ref{subsec:noise}} does not affect performance in clean conditions. The word error rates (WERs) in clean conditions are almost identical when using models trained with and without noise. Unsurprisingly, training with noise significantly improves performance in noisy conditions. For the \emph{Dictation} set, MTR training improves WER by around 60\% (relative). It is also interesting to note that MTR training only marginally affects performance of the 8 kHz test set.
\begin{table}[ht]
  \caption{Results using models trained w/ and w/o noise.}
  \label{tab:noise}
  \centering
  \begin{tabular}{lcrr}
    \toprule
    \multirow{2}{*}{\textbf{Test set}} & \multicolumn{1}{c}{\textbf{Sample}} & \multicolumn{2}{c}{\textbf{Word error rate}} \\
    {} & \multicolumn{1}{c}{\textbf{rate}} & \multicolumn{1}{c}{\textbf{w/o Noise}} & \multicolumn{1}{c}{\textbf{w/ Noise}} \\
    \midrule
    Voicesearch       & {16 kHz} & $10.5$ & $10.4$ \\
    {~~+~Noise}       & {16 kHz} & $17.7$ & $12.3$ \\
    Dictation         & {16 kHz} & $7.7$  & $7.8$ \\
    {~~+~Noise}       & {16 kHz} & $30.6$ & $12.3$ \\
    Dictation    & {8 kHz} & $8.2$  & $8.4$ \\
    \bottomrule
  \end{tabular}
\end{table}

\subsubsection{Codecs}
  {Tab.~\ref{tab:codec}} shows results when the models are trained with and without codec simulation. We evaluate on the \emph{Voicesearch} set under various codec conditions. MP3 at 64 kbps, Opus~\cite{opusRFC6716, ffmpeg} at 24 kbps, and SBC~\cite{Hoene2010SBC, SBCBluez} with a bitpool size of 24, 16 blocks and 8 subbands are unseen during training; the rest are seen. For the baseline model, performance worsens as the bitrate decreases since lower bitrates imply more lossy encoding. Training with codecs makes the performance under seen and unseen codecs close to not using any codec, even at low bitrates: For MP3 23k, training with codecs improves WER by almost 20\%.
  
\begin{table}[ht]
  \caption{Results using models trained with and without codec simulation. All of the results are on the Voicesearch test set.}
  \label{tab:codec}
  \centering
  \begin{tabular}{lrr}
    \toprule
    \multirow{2}{*}{\textbf{Test set}} & \multicolumn{2}{c}{\textbf{Word error rate}} \\
    {} & \textbf{w/o codec} & \textbf{w/ codec} \\
    \midrule
    No codec  & $10.5$ & $10.0$ \\
    AAC 23k   & $11.8$ & $11.4$ \\
    AAC 64k   & $10.6$ & $10.0$ \\
    MP3 23k   & $13.6$ & $10.6$ \\
    MP3 64k   & $10.5$ & $10.2$ \\
    OPUS 24k  & $10.8$ & $10.2$ \\
    SBC BLUEZ & $10.7$ & $10.2$ \\
    \bottomrule
  \end{tabular}
\end{table}

\subsubsection{Application domains}
  Next, we look at performance of application domain specific models. Results are shown in {Tab.~\ref{tab:domain}}. We present results using 3 domain-specific models trained on \emph{Voicesearch} and \emph{Dictation}, \emph{YouTube}, and \emph{Call-Center} data, respectively. Unsurprisingly, the models work well when the test domains match the model domains, and poorly when the domain changes. The \emph{YouTube} model, which is trained with the largest dataset among the 3 models, generalizes better. This is most likely because \emph{YouTube} training set includes data from a varied set of sources.
  
  The results also highlight the issues of training using a single domain, as is typically done in ASR. For example, training only on \emph{Call-center} data results in poor performance for test sets that are sampled at {16~kHz}. This is because \emph{Call-center} data is at {8~kHz}, and the model never sees utterances at a higher sampling rate during training. The model performs much better on \emph{Dictation} set when it is downsampled to {8~kHz}. On the \emph{Call-center} test set, the \emph{YouTube} model works as well as the domain-specific model. This is partly because the \emph{Call-center} training set is smaller compared to the rest of the sets, thereby limiting the generalization ability of the model trained with it.
  
  It is also interesting to see that all models perform poorly on the \emph{Telephony} test set, which is an unseen application domain for these models. As with the other test sets, the \emph{YouTube} model generalizes better and performs the best on this set.
  
\begin{table*}[ht]
  \caption{Results using domain-specific models.}
  \label{tab:domain}
  \centering
  \begin{tabular}{lcrrr}
    \toprule
    \multirow{2}{*}{\textbf{Test set}} & \multicolumn{1}{c}{\textbf{Sample}} & \multicolumn{3}{c}{\textbf{Word error rate}} \\
    {} & \multicolumn{1}{c}{\textbf{rate}} & \multicolumn{1}{c}{\textbf{Voicesearch-}} & \multicolumn{1}{c}{\textbf{Call-}}  & \multirow{2}{*}{\textbf{YouTube}} \\
    {} & {} & \multicolumn{1}{c}{\textbf{Dictation}} & \multicolumn{1}{c}{\textbf{center}} & {} \\
    \midrule
    Voicesearch & {16 kHz} & $10.5$ & $98.6$ & $16.3$ \\
    Dictation   & {16 kHz} & $7.7$  & $97.8$ & $13.9$ \\
    Dictation   & {8 kHz} & $8.2$  & $24.1$ & $13.9$ \\
    Call-center & {8 kHz} & $25.4$ & $20.5$ & $20.4$ \\
    YouTube     & {16 kHz} & $54.9$ & $97.2$ & $15.9$ \\
    Telephony   & {8 kHz} & $31.4$ & $27.8$ & $24.2$ \\
    \bottomrule
  \end{tabular}
\end{table*}

\subsubsection{Multidomain models}\label{subsec:multidomain}
  In {Tab.~\ref{tab:multidomain}}, we show results when the acoustic model is trained with all of the training data. Results are also shown with noise and codec simulation, in addition to multidomain training.
  
  Comparing these results with those in {Tab.~\ref{tab:domain}}, we can see that the multidomain model works similar to or better than the domain-specific models in all cases. It also generalizes better: It has better performance on the \emph{Telephony} set compared to the domain-specific models, and also on the noisy test sets compared to the baseline model trained without noise simulation. Finally, on the \emph{Call-center} set, it outperforms the domain-specific model, which shows that multidomain training partly addresses data sparsity issues. 
  
  When doing noise simulation, we selectively add noise to only those domains for which the input data is relatively clean. Specifically, we add noise only to \emph{Voicesearch}, \emph{Dictation}, and \emph{Call-center}. When the multidomain model is trained with noise simulation, it performs better on the noisy sets but degrades on the clean sets. Moreover, the performance in noisy conditions doesn't match those of the domain-specific model trained with noise ({Tab.~\ref{tab:noise}}). This is likely because the model doesn't see as much clean and noisy data as the domain-specific models because of the imbalance in training data distribution. Adding noise also makes the underlying task harder since the model has to learn to normalize out variations in the training data along several dimensions like application domain, noise and sample rate. It is interesting to see that even though large scale training works in some conditions, combinations of conditions makes it harder for the model to learn without any additional signal during training. As with noise, the multidomain model with codec simulation shows degradation on clean and noisy tests without codecs, as well as for the AAC 23k condition. It works better only on MP3 23k condition. In contrast, the domain-specific codec model performed better than its non-codec counterpart all around.
  
\begin{table*}[ht]
  \caption{Results using multidomain models.}
  \label{tab:multidomain}
  \centering
  \begin{tabular}{lcrrr}
    \toprule
    \multirow{2}{*}{\textbf{Test set}} &  \multicolumn{1}{c}{\textbf{Sample}} & \multicolumn{3}{c}{\textbf{Word error rate}} \\
    {} & \multicolumn{1}{c}{\textbf{rate}} & \multicolumn{1}{c}{\textbf{Multidomain}} & \multicolumn{1}{c}{\textbf{Multidomain}} & \multicolumn{1}{c}{\textbf{Multidomain}} \\
    {} & {} &  {} & \multicolumn{1}{c}{\textbf{+ Noise}} & \multicolumn{1}{c}{\textbf{+ Codec}} \\
    \midrule
    Voicesearch     & {16 kHz} & $10.2$ & $11.2$ & $10.5$   \\
    {~~+~AAC 23k}   & {16 kHz} & $11.4$ & $11.7$ & $11.9$    \\
    {~~+~MP3 23k}   & {16 kHz} & $13.5$ & $14.5$ & $11.1$   \\
    {~~+~Noise}     & {16 kHz} & $15.6$ & $13.9$ & $15.8$    \\
    Dictation       & {16 kHz} & $7.6$  & $8.4$  & $7.8$    \\
    {~~+~Noise}     & {16 kHz} & $24.5$ & $15.5$ & $24.3$    \\
    Dictation       & {8 kHz}  & $7.9$  & $10.1$ & $8.5$    \\
    Call-center     & {8 kHz}  & $16.4$ & $17.5$ & $17.6$    \\
    YouTube         & {16 kHz} & $16.3$ & $16.3$ & $16.6$    \\
    Telephony       & {8 kHz}  & $20.9$ & $22.3$ & $22.8$    \\
    \bottomrule
  \end{tabular}
\end{table*}
  
\subsection{Adaptation using multidomain models}\label{subsec:adaptation}
  One of the most important advantages of multidomain modeling is that it allows for easy and quicker adaptation to new conditions. To demonstrate this, we use varying amounts of in-domain data to improve performance on the \emph{Telephony} domain, which was not used to train the models in Sec.~\ref{subsec:multidomain}. We experiment with 10 hours, 30 hours, 100 hours and 700 hours (approx.) of training data. For adaptation, we use the relatively simpler fine-tuning approach. All layers of the model are tuned. A lower learning rate and early stopping is used to prevent over-fitting. We also train a separate model using the 700 hours training set, with model parameters initialized from scratch. Results are shown in {Tab.~\ref{tab:adapt}}.

\begin{table}[ht]
  \caption{Results on the Telephony test set when adapting the multidomain model and the Voicesearch-Dictation model using in-domain data. Also shown are the corresponding baselines from Tab.~\ref{tab:multidomain}, and when using a model trained only on Telephony training data.}
  \label{tab:adapt}
  \centering
  \begin{tabular}{lr}
    \toprule
    {\textbf{Test set}} & {\textbf{Word error rate}} \\
    \midrule
    Multidomain & $20.9$   \\
    + 10 hrs    & $13.2$ \\
    + 30 hrs    & $12.2$ \\
    + 100 hrs   & $11.3$ \\
    + 700 hrs   & $10.1$ \\
    \midrule
    Voicesearch- & \multirow{2}{*}{$31.4$} \\
    Dictation   & {} \\
    + 10 hrs    & $16.3$ \\
    + 30 hrs    & $14.6$ \\
    + 100 hrs   & $13.0$ \\
    + 700 hrs   & $10.9$ \\
    \midrule
    700 hrs     & $13.5$ \\
    \bottomrule
  \end{tabular}
\end{table}

  As can be seen, even though the multidomain model generalizes well compared to other domain specific models, it is still much worse than a model trained exclusively on \emph{Telephony} training set. But the multidomain model fine-tuned on just 10 hours of \emph{Telephony} speech works as well as a model trained on 700 hours from random initialization. Using more data for fine-tuning further improves performance. For example, the model fine-tuned using 100 hours of data outperforms the randomly initialized model by 16\% relative. And when using the entire 700 hours of data for fine-tuning, WER is better by 25\% relative compared to the randomly initialized model. 
  
  The table also shows results when the \emph{Voicesearch-Dictation} model is used for adaptation, instead of the multidomain model. As shown, as more and more in-domain data is used for adaptation, the difference compared to using the multidomain model for initialization decreases. But even with 700 hours of in-domain data, using the multidomain model works better by about 8\% relative. And with just 10 hours of data, adapting from the multidomain model is better by about 24\% relative. The \emph{Voicesearch-Dictation} model needs around 100 hours of in-domain data to reach a similar level of performance as the \emph{Telephony} model trained from scratch.
  
  The results clearly show that multidomain training allows for rapid adaptation. While fine-tuning to a particular subset this way comes at an expense of deteriorated WERs on the other subsets, we have noticed, in experiments not reported here, that by merging the \emph{Telephony} training set with mutlidomain training data, as in Sec.~\ref{subsec:multidomain}, helps retain the best performance in the remaining subsets. It is also possible to introduce domain specific parameters, as in \cite{khechai2018adaptation}, to avoid performance degradation when fine-tuning to a certain subset.
  
\section{Conclusion}\label{sec:concl}

 We presented a large scale study on domain robustness, training a single model by combining multiple application domains, sample rates, noise and codec conditions. Our results show that training at scale works well to address the domain mismatch problem: A model trained with data from multiple application domains worked as well as the domain-specific models. The multidomain model also showed better generalization properties, working better than domain-specific models for unseen application domains and in noisy conditions. Our results also show that multidomain models can be used to rapidly adapt to previously unseen conditions. 

 Multidomain training has several practical applications when deploying recognizers in practice, especially when it is hard to predict how the system will get used in the end. Having a single system also improves ease of maintainability. The ability to adapt it with small amounts of data makes it ideal to building task-specific models.
 
 One of the challenges going forward is the degradation in performance when training the model with multiple \emph{simulated} perturbation techniques. Future work will address this, by systematically dealing with the imbalance in training data distribution, especially when using simulation techniques. We also expect modeling strategies like domain adversarial training \cite{ganin2016domain} and better training techniques \cite{Shankar2018crossgrad} to improve performance. It will also be interesting to understand how to choose right subsets of data to be transcribed to help the model generalize better, and how to make use of unlabeled data that is available in much larger amounts compared to the labeled sets used in this work. 

\bibliographystyle{IEEEbib}

\begin{thebibliography}{10}

\bibitem{Stolcke2017human}
Andreas Stolcke and Jasha Droppo,
\newblock ``Comparing human and machine errors in conversational speech
  transcription,''
\newblock in {\em Proc. INTERSPEECH}, 2017, pp. 137--141.

\bibitem{Saon2017swbd}
George Saon, Gakuto Kurata, Tom Sercu, Kartik Audhkhasi, Samuel Thomas,
  Dimitrios Dimitriadis, Xiaodong Cui, Bhuvana Ramabhadran, Michael Picheny,
  Lynn-Li Lim, Bergul Roomi, and Phil Hall,
\newblock ``English conversational telephone speech recognition by humans and
  machines,''
\newblock in {\em Proc. INTERSPEECH}, 2017, pp. 132--136.

\bibitem{kim2017mtr}
Chanwoo Kim, Ananya Misra, Kean Chin, Thad Hughes, Arun Narayanan, Tara
  Sainath, and Michiel Bacchiani,
\newblock ``Generation of large-scale simulated utterances in virtual rooms to
  train deep-neural networks for far-field speech recognition in {Google
  Home},''
\newblock in {\em Proc. INTERSPEECH}, 2017.

\bibitem{peddinti2016mtr}
Vijayaditya Peddinti, Vimal Manohar, Yiming Wang, Daniel Povey, and Sanjeev
  Khudanpur,
\newblock ``Far-field {ASR} without parallel data.,''
\newblock in {\em Proc. INTERSPEECH}, 2016, pp. 1996--2000.

\bibitem{Yu2013FeatureLearningDNN}
D.~Yu, M.~L. Seltzer, J.~Li, J.-T. Huang, and F.~Seide,
\newblock ``Feature learning in deep neural networks - studies on speech
  recognition tasks,''
\newblock in {\em Proceedings of the International Conference on Learning
  Representations}, 2013.

\bibitem{Narayanan2013DNNSeparation}
A.~Narayanan and D.~L. Wang,
\newblock ``Investigation of speech separation as a front-end for noise robust
  speech recognition,''
\newblock {\em IEEE/ACM Transactions on Audio, Speech, and Language
  Processing}, vol. 22, pp. 826--835, 2014.

\bibitem{ghahremani17-transfer}
P.~Ghahremani, V.~Manohar, H.~Hadian, D.~Povey, and S.~Khudanpur,
\newblock ``Investigation of transfer learning for {ASR} using {LF-MMI} trained
  neural networks,''
\newblock in {\em Proc. ASRU}, 2017.

\bibitem{sainath2017multichannel}
Tara~N Sainath, Ron~J Weiss, Kevin~W Wilson, Bo~Li, Arun Narayanan, Ehsan
  Variani, Michiel Bacchiani, Izhak Shafran, Andrew Senior, Kean Chin, et~al.,
\newblock ``Multichannel signal processing with deep neural networks for
  automatic speech recognition,''
\newblock {\em IEEE/ACM Transactions on Audio, Speech, and Language
  Processing}, vol. 25, no. 5, pp. 965--979, 2017.

\bibitem{Narayanan2013IRM}
A.~Narayanan and D.~L. Wang,
\newblock ``Ideal ratio mask estimation using deep neural networks for robust
  speech recognition,''
\newblock in {\em Proceedings of the IEEE International Conference on
  Acoustics, Speech, and Signal Processing}, 2013, pp. 7092--7096.

\bibitem{yoshioka2015ntt}
Takuya Yoshioka, Nobutaka Ito, Marc Delcroix, Atsunori Ogawa, Keisuke
  Kinoshita, Masakiyo Fujimoto, Chengzhu Yu, Wojciech~J Fabian, Miquel Espi,
  Takuya Higuchi, et~al.,
\newblock ``The {NTT CHiME-3} system: Advances in speech enhancement and
  recognition for mobile multi-microphone devices,''
\newblock in {\em Automatic Speech Recognition and Understanding (ASRU), 2015
  IEEE Workshop on}. IEEE, 2015, pp. 436--443.

\bibitem{mirsamadi2017multi}
Seyedmahdad Mirsamadi and John~HL Hansen,
\newblock ``On multi-domain training and adaptation of end-to-end rnn acoustic
  models for distant speech recognition,''
\newblock in {\em Proc. INTERSPEECH}, 2017, pp. 404--408.

\bibitem{Li2017domain}
Jinyu Li, Michael~L. Seltzer, Xi~Wang, Rui Zhao, and Yifan Gong,
\newblock ``Large-scale domain adaptation via teacher-student learning,''
\newblock in {\em Proc. INTERSPEECH}, 2017, pp. 2386--2390.

\bibitem{gao2016experimental}
Jianqing Gao, Jun Du, Changqing Kong, Huaifang Lu, Enhong Chen, and Chin-Hui
  Lee,
\newblock ``An experimental study on joint modeling of mixed-bandwidth data via
  deep neural networks for robust speech recognition,''
\newblock in {\em Neural Networks (IJCNN), 2016 International Joint Conference
  on}. IEEE, 2016, pp. 588--594.

\bibitem{bengio2012transfer}
Yoshua Bengio,
\newblock ``Deep learning of representations for unsupervised and transfer
  learning,''
\newblock in {\em Proceedings of ICML Workshop on Unsupervised and Transfer
  Learning}, 2012, pp. 17--36.

\bibitem{switchboard}
John~J Godfrey, Edward~C Holliman, and Jane McDaniel,
\newblock ``Switchboard: Telephone speech corpus for research and
  development,''
\newblock in {\em Acoustics, Speech, and Signal Processing, 1992. ICASSP-92.,
  1992 IEEE International Conference on}. IEEE, 1992, vol.~1, pp. 517--520.

\bibitem{Paul1992}
D.~Paul and J.~Baker,
\newblock ``The design of {W}all street journal-based {CSR} corpus,''
\newblock in {\em Proceedings of the International Conference on Spoken
  Language Processing}, 1992, pp. 899--902.

\bibitem{ami-corpus}
Iain McCowan, Jean Carletta, W~Kraaij, S~Ashby, S~Bourban, M~Flynn,
  M~Guillemot, T~Hain, J~Kadlec, V~Karaiskos, et~al.,
\newblock ``The ami meeting corpus,''
\newblock in {\em Proceedings of the 5th International Conference on Methods
  and Techniques in Behavioral Research}, 2005, vol.~88, p. 100.

\bibitem{ganin2016domain}
Yaroslav Ganin, Evgeniya Ustinova, Hana Ajakan, Pascal Germain, Hugo
  Larochelle, Fran{\c{c}}ois Laviolette, Mario Marchand, and Victor Lempitsky,
\newblock ``Domain-adversarial training of neural networks,''
\newblock {\em The Journal of Machine Learning Research}, vol. 17, no. 1, pp.
  2096--2030, 2016.

\bibitem{tzeng2017adversarial}
Eric Tzeng, Judy Hoffman, Kate Saenko, and Trevor Darrell,
\newblock ``Adversarial discriminative domain adaptation,''
\newblock in {\em Computer Vision and Pattern Recognition (CVPR)}, 2017,
  vol.~1, p.~4.

\bibitem{brandenburg1999mp3}
Karlheinz Brandenburg,
\newblock ``{MP3} and {AAC} explained,''
\newblock in {\em Audio Engineering Society Conference: 17th International
  Conference: High-Quality Audio Coding}, Sep 1999.

\bibitem{ffmpeg}
``{FFmpeg},''
 \newblock{{www.ffmpeg.org}} .

\bibitem{pundak2016lfr}
Golan Pundak and Tara~N Sainath,
\newblock ``Lower frame rate neural network acoustic models.,''
\newblock in {\em Proc. INTERSPEECH}, 2016, pp. 22--26.

\bibitem{sak2014LSTM}
H.~Sak, A.~Senior, and F.~Beaufays,
\newblock ``Long short-term memory recurrent neural network architectures for
  large scale acoustic modeling,''
\newblock in {\em Proc. INTERSPEECH}, 2014.

\bibitem{allauzen2011bayesian}
Cyril Allauzen and Michael Riley,
\newblock ``Bayesian language model interpolation for mobile speech input,''
\newblock in {\em Proc. Interspeech}, 2011.

\bibitem{abadi2016tensorflow}
Mart{\'\i}n Abadi, Paul Barham, Jianmin Chen, Zhifeng Chen, Andy Davis, Jeffrey
  Dean, Matthieu Devin, Sanjay Ghemawat, Geoffrey Irving, Michael Isard,
  et~al.,
\newblock ``Tensorflow: a system for large-scale machine learning.,''
\newblock in {\em OSDI}, 2016, vol.~16, pp. 265--283.

\bibitem{variani2017end}
Ehsan Variani, Tom Bagby, Erik McDermott, and Michiel Bacchiani,
\newblock ``End-to-end training of acoustic models for large vocabulary
  continuous speech recognition with tensorflow,''
\newblock in {\em Proc. Interspeech}, 2017, pp. 1641--1645.

\bibitem{jouppi2017datacenter}
Norman~P Jouppi, Cliff Young, Nishant Patil, David Patterson, Gaurav Agrawal,
  Raminder Bajwa, Sarah Bates, Suresh Bhatia, Nan Boden, Al~Borchers, et~al.,
\newblock ``In-datacenter performance analysis of a tensor processing unit,''
\newblock in {\em Computer Architecture (ISCA), 2017 ACM/IEEE 44th Annual
  International Symposium on}. IEEE, 2017, pp. 1--12.

\bibitem{liao2013large}
Hank Liao, Erik McDermott, and Andrew Senior,
\newblock ``Large scale deep neural network acoustic modeling with
  semi-supervised training data for youtube video transcription,''
\newblock in {\em Automatic Speech Recognition and Understanding (ASRU), 2013
  IEEE Workshop on}. IEEE, 2013, pp. 368--373.

\bibitem{soltau2016neural}
Hagen Soltau, Hank Liao, and Hasim Sak,
\newblock ``Neural speech recognizer: Acoustic-to-word {LSTM} model for large
  vocabulary speech recognition,''
\newblock {\em arXiv preprint arXiv:1610.09975}, 2016.

\bibitem{Senior2014}
Andrew Senior and Ignacio Lopez-Moreno,
\newblock ``Improving {DNN} speaker independence with i-vector inputs,''
\newblock in {\em Acoustics, Speech and Signal Processing (ICASSP), 2014 IEEE
  International Conference on}. IEEE, 2014, pp. 225--229.

\bibitem{rousseeuw1987}
Peter~J. Rousseeuw,
\newblock ``Silhouettes: A graphical aid to the interpretation and validation
  of cluster analysis,''
\newblock {\em Computational and Applied Mathematics}, vol. 20, pp. 53--65,
  1987.

\bibitem{Davies1979}
David~L. Davies and Donald~W. Bouldin,
\newblock ``A cluster separation measure,''
\newblock {\em {IEEE} Transactions on Pattern Analysis and Machine
  Intelligence}, vol. 1, pp. 224--227, April 1979.

\bibitem{opusRFC6716}
JM. Valin, K.~Vos, and T.~Terriberry,
\newblock ``Definition of the {Opus Audio Codec},''
\newblock RFC 6716, RFC Editor, September 2012.

\bibitem{Hoene2010SBC}
C.~Hoene and M.~Hyder,
\newblock ``Optimally using the bluetooth subband codec,''
\newblock in {\em {Proceedings of the IEEE 35th Conference on Local Computer
  Networks}}, 2010, pp. 356--359.

\bibitem{SBCBluez}
``{SBC} library,'' {O}nline: www.bluez.org.

\bibitem{khechai2018adaptation}
Khe~Chai Sim, Arun Narayanan, Ananya Misra, Anshuman Tripathi, Golan Pundak,
  Tara~N. Sainath, Parisa Haghani, Bo~Li, and Michiel Bacchiani,
\newblock ``Domain adaptation using factorized hidden layer for robust
  automaticspeech recognition,''
\newblock in {\em Proc. INTERSPEECH}, 2018.

\bibitem{Shankar2018crossgrad}
S.~Shankar, V.~Piratla, S.~Chakrabarti, S.~Chaudhuri, P.~Jyothi, and
  S.~Sarawagi,
\newblock ``Generalizing across domains via cross-gradient training,''
\newblock in {\em Proceedings of the International Conference on Learning
  Representations}, 2018.

\end{thebibliography}

\end{document}